%% file: main.tex
\documentclass[10pt,twocolumn,letterpaper]{article}

\usepackage[pagenumbers]{cvpr} 

\input{preamble}

\definecolor{cvprblue}{rgb}{0.21,0.49,0.74}
\usepackage[pagebackref,breaklinks,colorlinks,allcolors=cvprblue]{hyperref}
\usepackage{algorithm, algorithmicx, algpseudocode}
\usepackage{tabularray}
\usepackage{tcolorbox}
\usepackage{adjustbox}
\usepackage{fancyvrb}
\usepackage{hyperref}
\usepackage[accsupp]{axessibility}

\title{Global-Local Tree Search in VLMs for 3D Indoor Scene Generation}

\author{
Wei Deng \qquad 
Mengshi Qi\textsuperscript{\thanks{Corresponding author: qms@bupt.edu.cn}} \qquad 
Huadong Ma \\
State Key Laboratory of Networking and Switching Technology, \\ Beijing University of Posts and Telecommunications, China \\
\textit{\{dw-dengwei, qms, mhd\}@bupt.edu.cn}
}

\begin{document}
\maketitle

\input{sec/0_abstract}    
\input{sec/1_intro}
\input{sec/2_related}

\input{sec/3_method}
\input{sec/4_exp}
\input{sec/5_conclusion}
\input{sec/6_ack}

\nocite{SGFormer, lv_neurips, qi_tip, STC-GAN, qi_CVPR19, qi_tcsvt, RDFC-GAN}
{
    \small
    \bibliographystyle{ieeenat_fullname_etal}
    \bibliography{main}
}

\end{document}

%% file: preamble.tex
\makeatletter
\DeclareRobustCommand\onedot{\futurelet\@let@token\@onedot}
\def\@onedot{\ifx\@let@token.\else.\null\fi\xspace}
\def\eg{\emph{e.g}\onedot} 
\def\ie{\emph{i.e}\onedot}

\def\etc{\emph{etc}\onedot}

\def\etal{\emph{et al}\onedot}
\def\versus{\emph{vs}\onedot}
\makeatother

%% file: sec/0_abstract.tex
\begin{abstract}
Large Vision-Language Models (VLMs), such as GPT-4, have achieved remarkable success across various fields. 
However, there are few studies on 3D indoor scene generation with VLMs. 
This paper considers this task as a planning problem subject to spatial and layout common sense constraints.
To solve the problem with a VLM, we propose a new global-local tree search algorithm.
Globally, the method places each object sequentially and explores multiple placements during each placement process, where the problem space is represented as a tree.
To reduce the depth of the tree, we decompose the scene structure hierarchically, \ie room level, region level, floor object level, and supported object level.
The algorithm independently generates the floor objects in different regions and supported objects placed on different floor objects.
Locally, we also decompose the sub-task, the placement of each object, into multiple steps.
The algorithm searches the tree of problem space.
To leverage the VLM model to produce positions of objects, we discretize the top-down view space as a dense grid and fill each cell with diverse emojis to make to cells distinct.
We prompt the VLM with the emoji grid and the VLM produces a reasonable location for the object by describing the position with the name of emojis.
The quantitative and qualitative experimental results illustrate our approach generates more plausible 3D scenes than state-of-the-art approaches.
Our source code is available at \href{https://github.com/dw-dengwei/TreeSearchGen}{https://github.com/dw-dengwei/TreeSearchGen} .

\end{abstract}

%% file: sec/1_intro.tex
\section{Introduction}
\label{sec:intro}
\begin{figure}
    \centering
    \includegraphics[width=.9\linewidth]{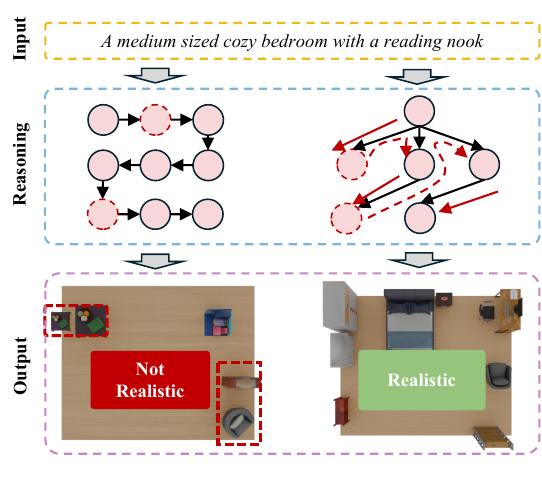}
    \caption{
    Illustration of chain-like (left) and tree-like (right) reasoning in VLMs on 3D scene generation.
    Each node represents a token or language sequence.
    The red dashed nodes indicate the VLM produces an inappropriate output.
    The chain-like method cannot correct the prior errors and the subsequent process reasons based on the errors, leading to a non-realistic layout, such as exceeding the room.
    In contrast, the tree-like method can modify the output if a mistake occurs, resulting in a more realistic layout.
    }
    \label{fig:intro-fig}
\end{figure}
3D indoor scene generation~\cite{AnyHome,ATISS,CommonScenes,DiffuScene,Graph-to-3d,InstructScene,LayoutGPT,PhyScene,SceneHGN,SceneFormer,HoloDeck} refers to the process of automatically producing realistic 3D indoor scenes with a computer program.
In recent years, some studies~\cite{InstructScene,LayoutGPT,HoloDeck,AnyHome,DiffuScene} have explored on using natural language for scene generation, as the textual prompts offer a user-friendly interface.
Realistic 3D indoor scene generation is crucial in fields such as interior design, 3D gaming, virtual/augmented reality, and embodied AI.

The primary challenge in generating high-quality scenes is modeling plausible and realistic spatial relationships (\ie, 3D layouts) between objects.
To achieve this, early studies~\cite{InstructScene,DiffuScene} adopt a data-driven fashion.
They train generative models on 3D scene datasets to learn the joint distribution of real indoor scenes and their textual descriptions.
However, collecting 3D scene datasets is challenging and costly, resulting in limited dataset scales and less robust models.
Recently, researchers have focused on using large Vision-Language Models (VLMs)~\cite{AnyHome,LayoutGPT,HoloDeck,GALA3D} to achieve this task.
VLMs are trained on large-scale datasets, enabling them to understand complex user instructions and infer 3D layouts from language based on common knowledge of indoor furnishing.
For example, Feng~\etal~\cite{LayoutGPT} propose LayoutGPT to generate the dimension, location, and orientation parameters for each object within a scene from a textual prompt.
However, as shown in~\Cref{fig:intro-fig}, the current VLMs perform a token-level, left-to-right decision-making process during inference time~\cite{ToT}, which is not suited for 3D layout reasoning. 
Specifically, such an auto-regressive reasoning method cannot modify previous outputs.
If a previous object is placed at an inappropriate location, this error will accumulate.
Consequently, it remains a challenge to improve VLM's reasoning ability for 3D scene generation.

3D scene generation needs to search with a tree-like approach rather than chain-like.
When furnishing a room, humans usually place objects sequentially.
In addition, each object has multiple candidate positions.
We pick one of them for the current object and continue to put the next objects.
This process is iteratively applied until we place all the objects.
If an object is put in an inappropriate location, it stops the next object from being put in the room and we will adjust the previous decision.
Consequently, in~\Cref{fig:intro-fig}, the problem space can be represented as a tree.
Each layer in the tree denotes an individual object and each node in the same layer represents the candidate placement for the object.
The chain from the root node to the last layer is a layout solution.
Our objective is to search on the tree to find such a chain, subjected to space range, the placement common sense, non-overlapping, and non-floating constraints.

Based on the above analysis, we propose a novel global-local tree search method to enhance VLM reasoning for 3D indoor scene generation.
Complex scenes with numerous objects make searching a deep tree challenging.
Our key insight is that the scene structures can be represented hierarchically.
We first construct a hierarchical scene representation from user input.
A scene is decomposed as room level, region level, floor object level, and supported object level.
This representation serves as a proxy between the textual input and the scene output and we can place the objects region by region, which reduces the computational cost.
Subsequently, we propose a global-local tree search method to generate the layout for the object within the same region.
The global tree search method places objects sequentially, mimicking human behavior.
Specifically, it starts from the root node (the region) and leverages the local tree search method to generate a node in the next layer.
Then, the global tree search method turns to the next layer and iteratively applies such a process until it generates all objects.
The local tree search method generates the position of an individual object.
At each layer, it determines one position parameter by textually and visually prompting the VLM.
In our approach, each layer has multiple alternatives.
Thus, our method performs a tree search in the problem space.

To summarize, our contributions are three-fold:

\noindent (1) We propose a novel global-local tree search method to enhance VLM reasoning for generating realistic 3D indoor scenes;

\noindent (2) We design a hierarchical scene representation as a commonsense bridge between textual input and 3D indoor scenes, further reducing computational costs.

\noindent (3) Our quantitative and qualitative studies demonstrate that our method generates more realistic 3D indoor scenes than state-of-the-art approaches. User studies suggest that our approach ranks the best among the three approaches.

%% file: sec/2_related.tex
\section{Related Work}
\label{sec:related}

\textbf{3D Scene Generation} is not as trivial as directly using a 3D object generation model to generate multiple objects.
The primary challenge lies in layout generation, which requires modeling spatial and semantic relationships among objects.
In the early stage, some works train generative models (\eg GANs~\cite{Giraffe, BlockGAN}, VAEs~\cite{Graph-to-3d, CommonScenes, SceneHGN}, diffusion models~\cite{CommonScenes, DiffuScene, InstructScene, PhyScene}, and auto-regressive models~\cite{ATISS, SceneFormer}) with 3D scene datasets~\cite{3D-FRONT, 3D-FUTURE, 3DSSG}.
However, the 3D scene datasets are considerably small compared with 3D object datasets (3D-FRONT~\cite{3D-FRONT} 18k+ \versus Objaverse~\cite{objaverse} 800k+ and Objaverse-XL~\cite{Objaverse-XL} 10M+).
Thus, models trained on small-scale datasets are less robust and limited in achieving satisfactory performance.
Recently, some works~\cite{CLIP,LayoutGPT,HoloDeck,AnyHome} utilize the commonsense of VLMs to understand user-provided instructions and generate layouts.
For example, they use CLIP~\cite{CLIP} to retrieve the most relevant 3D objects by measuring the similarity scores between the textual object descriptions and the image renderings of the objects from a 3D object database.
LayoutGPT~\cite{LayoutGPT} directly outputs the scene layout formatted as Cascading Style Sheets (CSS), including size, location, and orientation for each object.
However, the current pre-trained language models cannot comprehensively perceive the space, and thus usually yield unsatisfactory results, such as objects intersecting with each other.
To this end, HoloDeck~\cite{HoloDeck} and AnyHome~\cite{AnyHome} propose to generate scene graphs, which represent the objects as nodes and spatial relationships as edges, and propose rule-based algorithms to convert the scene graph into the room layout.
However, the rules are imperfect, leading to low diversity and impractical results.
In our work, we strive to utilize the VLM to perceive the space of indoor scenes and achieve reasonable object placements rather than a rule-based algorithm.

\noindent\textbf{Reasoning with Vision-Language Models} is very important in problem-solving, decision-making, and critical thinking~\cite{reason-survey}.
VLMs can be prompted with a standard input-output paradigm~\cite{GPT3, GPT4, PaLM,LLaMA,Qwen-VL,LLaVA}.
However, language models are trained to generate coherent language sequences.
Such a simple prompting method falls short when the task is complex and requires multi-step reasoning.
To enable step-by-step reasoning Wei~\etal~\cite{CoT} propose Chain-of-Thought (CoT), which enforces the language models output intermediate thoughts.
CoT significantly improves the performance on reasoning tasks.
Nevertheless, some tasks require exploring multiple alternatives at each intermediate step rather than just picking one.
To this end, Yao~\etal~\cite{ToT} propose Tree-of-Thoughts (ToT), which maintains a tree of thoughts and leverages classical tree search algorithms to find solutions for a complex task.
Inspired by ToT, we treat generating layouts for 3D scenes as a tree search problem and propose the global-local tree search method to generate layouts.

%% file: sec/3_method.tex
\section{Problem Formulation}\label{sec:formulation}
Given a textual prompt $x$, we aim to generate a 3D scene with $N$ objects $S=\{o_1, o_2, \cdots, o_N\}$ with a VLM, where each object $o_i=(c_i, s_i, p_i, r_i)$, $c_i$ is the category of the object, $s_i$, $p_i$, and $r_i$ denote size, position and orientation, respectively.
The standard input-output (IO) reasoning method directly produces a 3D scene: $S\sim p^{\mathrm{IO}}(S|x)$.
During the reasoning of VLMs, the chain-of-thought (CoT) method yields a chain of intermediate thoughts $t_1, t_2, \cdots, t_n$ to bridge $x$ and $S$.
Each intermediate thought $t_i\sim p^{\mathrm{CoT}}(t_i|x,t_1, t_2, \cdots, t_{i-1})$ is sampled auto-regressively and produces the final output $S\sim p^{\mathrm{CoT}}(S|x,t_1, t_2, \cdots, t_n)$.
However, there is a large semantic gap between the input and output, thus it is hard for the current method to directly map from $x$ to $S$.
Besides, both IO and CoT methods follow a left-to-right manner, which cannot modify previous decisions.

In our work, we first generate a proxy $P$ for linking the user input and the 3D scene: $p^{x\to S}=p^{P\to S}(S|P)p^{x\to P}(P|x)$.
It is a hierarchical representation, which decomposes a scene into multiple regions (see more details in~\Cref{sec:hierarchical}).
We independently generate all the regions and then combine the regions into a whole room.
Each region contains an object set $S'=\{o_1',o_2',\cdots,o_N'\}$ and an edge set $E=\{e_{ij}|o_i,o_j\in S\}$.
The object set is different from $S=\{o_1,o_2,\cdots,o_N\}$.
In $S'$, the category $c_i$ and size $s_i$ are determined while the placement attributes, \ie $p_i$ and $r_i$, have not been determined in $p^{x\to P}(P|x)$.
There is an anchor object in $S'$ that represents the primary function of this region.
Each edge is the spatial relationship between an object and the anchor, denoted as $e_{i,a}$.

In $p^{P\to S}(S|P)$, we propose an improved reasoning method as a tree search for 3D scene generation (see more details in ~\Cref{sec:global-local}).
We aim to determine the orientation $r_i$ and location $p_i$ for all $o_i'\in S'$.
We propose a global-local tree search method to achieve this.
It searches on the problem space, a tree, aiming at finding a solution for the placements of each object, and will trace back if it fails to go deeper into the tree.
The global tree search method is an object-level solver, which manages the generating process globally.
It puts the objects via the local tree search method sequentially as the human does: $o_{i+1}\sim p(o_{i+1}|e_{ia},o_1, o_2, \cdots, o_{i}, o_{i+1}')$.
It suggests that the determination of the orientation and location of the $(i+1)^{\mathrm{th}}$ object is conditioned on the objects whose orientation and location are determined, relationship with the anchor, and the category $c_{i+1}$ and size $s_{i+1}$ of the $(i+1)^{\mathrm{th}}$ object itself.
In contrast, the local tree search method is a parameter-level solver.
It decomposes the generation of the orientation and location into smaller sub-tasks.
Each sub-task also has multiple alternatives and the problem space spans as a tree.

In the following sections of this paper, we first introduce the hierarchical scene representation in~\Cref{sec:hierarchical}.
Then, we propose our global-local tree search method in~\Cref{sec:global} and~\Cref{sec:local}, respectively.

\section{Hierarchical Scene Representation}
\label{sec:hierarchical}
\begin{figure*}
    \centering
    \includegraphics[width=1.0\linewidth]{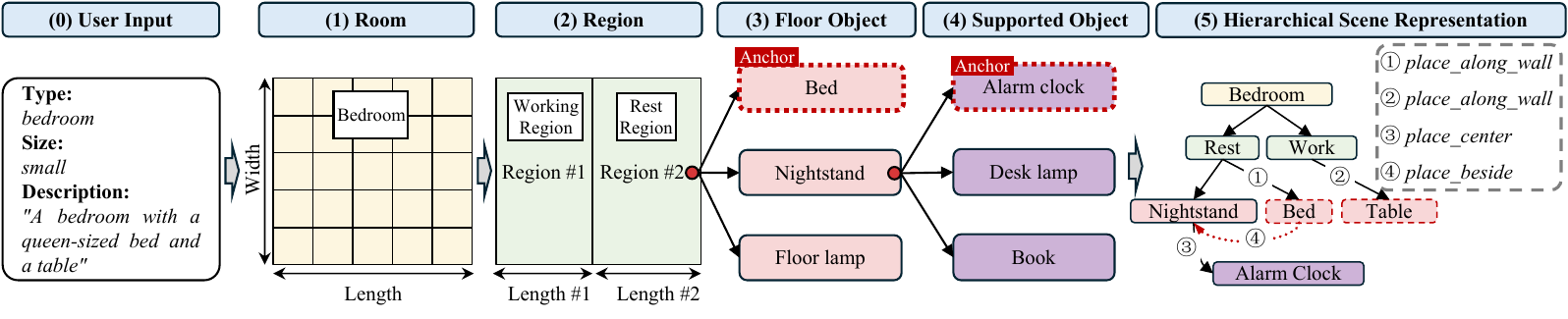}
    \caption{
    We prompt a VLM to generate the hierarchical scene representation level by level.
    From left to right, we decompose the scene into room, region, floor object, and supported object levels.
    The final representation is shown on the right-most side in this figure.
    }
    \label{fig:scene-graph}
\end{figure*}

Generating a 3D indoor scene from natural language is a complex task because there is a large semantic gap between the text and the scene.
To this end, we first leverage the VLM to understand the user requirements and produce a hierarchical scene structure representation, serving as a proxy between the text and the scene.
In this section, we introduce the hierarchical representation and how to generate such a representation with the VLM.

We represent the scene structure hierarchically including room, region, floor object, and supported object levels, as shown in~\Cref{fig:scene-graph}.
We start from the user input and prompt the VLM to generate the hierarchical scene representation from left to right.
During the generation process, we explicitly prompt the VLM to follow the user's input and consider the common sense of indoor furnishing.

\noindent\textbf{Room level.} In ~\Cref{fig:scene-graph} (1), the first level is the root node, denoting the entire room $P$. 
We prompt the VLM to generate a reasonable dimension for the room, parameterized by the length and width $\dim=(l,w)$.

\noindent\textbf{Region level.} In ~\Cref{fig:scene-graph} (2), the second level is the region level, we divide the room into multiple functional regions.
Each region is characterized by its dimensions $\dim_i=(l_i,w_i)$.
We enforce them to share the width with the room to ease the calculation of region positions and absolute object positions in the region.

\noindent\textbf{Floor object level.} In ~\Cref{fig:scene-graph} (3), the third level is the floor object level, which represents the objects that should be placed on the floor, called floor objects.
Each floor object should belong to exactly one region.
As described in~\Cref{sec:formulation}, we utilize the VLM to generate $S'$ and $E$ for each region.
We first prompt the VLM to generate some objects with its category $c_i$ and size $s_i$, where the object should be semantically consistent with the region function.
We request the VLM to choose an anchor object from the object set.
Other objects are spatially related to the anchor object, such as a coffee table placed in front of the sofa (anchor), and we also utilize the VLM to determine the spatial relationships, \ie edge between objects and the anchor $e_{i,a}$.
The VLM can choose from the following options:
\textit{place\_front}, \textit{place\_beside}, and \textit{place\_around}.

To get the 3D object models in the object set, we retrieve them from the Objaverse-1.0 database~\cite{objaverse}.
We adopt the process outlined in HoloDeck~\cite{HoloDeck}.
We use the CLIP model~\cite{CLIP} to measure visual similarity, Sentence-BERT for textual similarity~\cite{SentenceBERT}, and compute dimension discrepancies for object retrieving.

In particular, we use the VLM to generate the placement rule for the anchor object, \ie \textit{place\_along\_wall}, \textit{place\_in\_center}, and \textit{place\_at\_corner}.
The anchor always faces the free space of the region.
The anchor position and orientation are important because we need to use $e_{i,a}$ to reason the placement of other objects.

\noindent\textbf{Supported object level.} 
In ~\Cref{fig:scene-graph} (4), the fourth level is the supported object level, which presents the objects that should be placed on the floor objects, named supported objects.
Some floor objects, like desks and nightstands, can support other objects, while others, such as wardrobes and floor lamps, cannot.
Similarly, we also build an object set and an edge set and retrieve 3D models for the supported objects akin to the process at the floor object level.

\section{Global-Local Tree Search in VLM}
\label{sec:global-local}

\begin{figure*}
    \centering
    \includegraphics[width=1\linewidth]{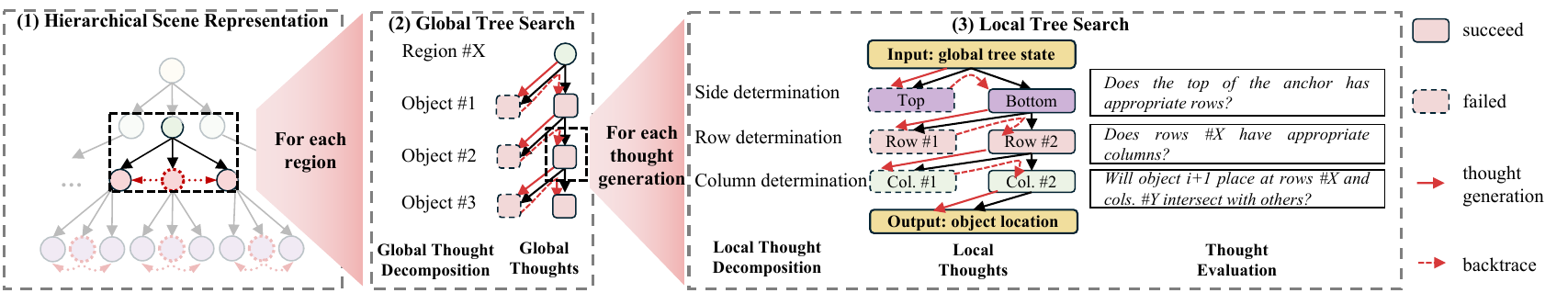}
    \caption{
    To generate a layout for a scene with quantities of objects, we independently generate the layout for each region.
    The global and local tree search method starts from the root node and goes deep by generating a thought.
    If the thought generator fails to produce a thought, it will trace back to the parent node and move to another thought.
    }
    \label{fig:global-local}
\end{figure*}

Based on the analysis in~\Cref{sec:formulation} and the proxy $P$ in~\Cref{sec:hierarchical}, we propose a global-local tree search method to reason a 3D scene with a VLM via $p^{P\to S}(S|P)$.

\begin{figure}
    \centering
    \includegraphics[width=1.\linewidth]{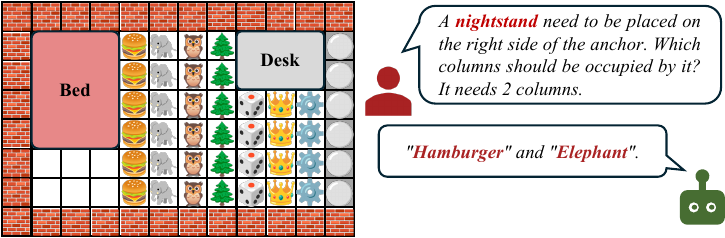}
    \caption{
    We discretize the top-down view as a grid and fill the cells with emojis.
    The \textit{brick} and \textit{white go} emojis stand for the wall and region boundary respectively.
    }
    \label{fig:emoji}
\end{figure}
To make the VLM reason spatially, which requires perceiving the scene, we feed textual-visual prompts to the VLM.
Specifically, we discretize the top-down view of a scene layout as a grid, as shown in~\Cref{fig:emoji}.
In the grid, the existing objects are represented as rectangles and the anchor object is highlighted in red.
We prompt the VLM with the grid and textual instructions, guiding it to reason spatially on the grid
For example, we wish to put a \textit{nightstand}, which occupies two columns, to the right side of the anchor.
The cells on the right side of the anchor are filled with distinct emojis to make the VLM distinguish the cells.
The VLM is asked to answer the name of the emojis of the columns where the \textit{nightstand} should be placed.

\subsection{Global Tree Search}\label{sec:global}
Given the object set $S'$ and the edge set $E$ defined in the hierarchical scene representation, we aim to reason the orientation and location of each object.

For orientations, we prompt the VLM to select a rule from the following options: \textit{face to anchor}, \textit{back to anchor}, \textit{face the same direction of the anchor}, and \textit{face the opposite direction of the anchor}.
Based on the rules and the anchor's location and orientation, we can determine the orientation of the non-anchor objects.

Based on the problem formulation in~\Cref{sec:formulation}, we treat it as a tree search problem.
To solve this, we propose our global tree search method to globally maintain this process.
The global tree search method can be described as:
(1) decompose the thoughts,
(2) generate the thoughts,
and (3) search the tree of thoughts to find a solution with the depth-first-search (DFS) algorithm, where thoughts stand for the nodes in the problem tree.

\noindent\textbf{Thought decomposition.}
As shown in~\Cref{fig:global-local} (2), the tree is started from the region node and each layer represents an individual object.
We decompose the task into the placement for each object.
We initially place the anchor object in the empty region.
The remaining objects are then placed in descending order of their dimensions.

\noindent\textbf{Thought generation.}
In~\Cref{fig:global-local} (2), given the relationship between the $(i+1)^{\mathrm{th}}$ object and the anchor $e_{i+1,a}$ and a tree state $s_i=\{o_1, o_2, \cdots, o_i\}$, we propose at most $k$ alternative thoughts for the $(i+1)^{\mathrm{th}}$ object.
That is, $o_{i+1}^{(j)} \sim p_\theta(o_{i+1}|s_{i},e_{i+1,a}) (j=1, \cdots, k)$, where $p_\theta$ is a thought generator.
The thought generator $p_\theta$ is the local tree search method, which will be introduced in~\Cref{sec:local}.

\noindent\textbf{Search algorithm.}
We aim to explore the most promising solution first, thus we use the DFS algorithm to search for the solution.
As shown in~\Cref{fig:global-local} (2), the algorithm starts from the region node.
The algorithm iteratively generates a thought in each layer with the local tree search method.
If it successfully generates a thought in layer $i+1$, \ie enough space to satisfy the spatial relation, the algorithm will walk to the next layer of the tree.
Otherwise, the algorithm will propose thoughts via the thought generator at most $k$ times $o_{i+1}^{(1\cdots k)}$ until succeeds.
If all the $k$ attempts fail, it suggests that the previous placements, $o_1, o_2, \cdots, o_i$, tend to be inappropriate.
The algorithm then traces back to layer $i$ and regards it fails in the layer.
Then, it will retry in layer $i$.
The algorithm performs iteratively and ends after successfully proposing a thought in the last layer of the tree.

\subsection{Local Tree Search}
\label{sec:local}

The local tree search method plays the thought generator for the global tree search method.
It takes as input the relation between the $(i+1)^{\mathrm{th}}$ object and the anchor $e_{i+1,a}$ and the intermediate tree state $s_i=[o_1, o_2, \cdots, o_i]$, and outputs the position $p_{i+1}$ of the $(i+1)^{\mathrm{th}}$ object.
We also regard location determination as a tree search problem, named local tree search.
The algorithm is described as follows:

\noindent\textbf{Thought decomposition.}
First, we determine the object should be placed on which side of the anchor in the top-down view (\ie left, right, top, and bottom).
Second, if the side determination step chooses \textit{top} or \textit{bottom}, we determine the rows in the grid as shown in~\Cref{fig:emoji} where the new object should be placed.
On the contrary, if the side is \textit{left} or \textit{right}, we determine columns in this step.
Third, we determine the axis which is different from Step 2.
At each step, there are also multiple alternatives.
Thus the problem space can be presented as a tree.

\noindent\textbf{Thought generation.}
At each step, we textually and visually prompt the VLM, as shown in~\Cref{fig:emoji}.
The VLM is prompted to consider the layout common sense to generate reasonable thoughts.
For side determination, the VLM should choose one of the side options.
For row and column determinations, the VLM needs to choose the emojis where the number of them is equal to the object dimension.

\noindent\textbf{Thought Evaluation.}
The VLM needs to evaluate intermediate thoughts.
For example, in the side determination step, we prompt the VLM to evaluate if the chosen side has an appropriate position to put the new object.
In particular, for the last step, we check whether the bounding box of the new object will intersect with others.
The output of the evaluator means whether the current step succeeds.

\noindent\textbf{Search algorithm.}
The local tree search shares the same search algorithm with the global counterpart. 
If the local tree search fails to produce all the thoughts in the three layers although it fully searches the tree with maximum $k$ attempts, it suggests the global tree search method unable to place the current object.

%% file: sec/4_exp.tex
\section{Experiments}
\input{tab/prompt}

In this section, we present our experimental results on 3D scene generation and provide comparisons with state-of-the-art approaches.
\label{sec:experiments}

\subsection{Setup}
\noindent\textbf{Input Prompts.}
We leverage ChatGPT to produce textual prompts for four types of scenes: \textit{bathroom}, \textit{bedroom}, \textit{kitchen}, and \textit{living room}.
Each prompt consists of \textit{room size} (one of \textit{small}, \textit{medium}, or \textit{large}) and a description of how to furnish the room, such as ``\textit{A room with a queen-sized bed for sleeping and an office table for working}''.
We generate 120 prompts, with 30 prompts for each scene type.
All the generated samples in our experiments originate from these prompts.
\Cref{tab:prompt} illustrates some example inputs in our experiment generated by ChatGPT.

\noindent\textbf{Parameter Settings.}
We apply OpenAI GPT-4o API to conduct our experiments.
The breadth (the maximum attempts $k$) of the tree is important for the global-local tree search algorithm.
If $k$ is too small, it may not successfully find an optimal solution.
If $k$ is too large, the search space will be large and the API cost will be extremely high.
Therefore, we make a trade-off between effect and cost.
In the global tree search module, we set $k=3$ for the anchor objects.
For other objects, we set $k=1$.
In the local tree search module, we set $k=2$ for the side determination step and set $k=1$ for others.

\noindent\textbf{Metrics.}
To quantitatively measure the methods, we follow HoloDeck~\cite{HoloDeck} to compute the \textbf{CLIP score}~\cite{CLIP}.
Specifically, we use the OpenCLIP library with the ViT-L/14 model pre-trained on the LAION-2B dataset~\cite{LAION-5B}.
Subsequently, we calculate the cosine similarity between the top-down rendering of scenes and a prompt template ``\textit{a top-down view of [scene type]}".
We multiply the similarities by 100 as the CLIP score.

To qualitatively measure the methods, we calculate the \textbf{reciprocal rank}.
We first render the scenes from the top-down view for each scene with Blender.
Subsequently, we pair the results from different methods feeding the same input and shuffle them.
We invite 15 annotators to rank the generated scenes.
They are asked \textit{``Which scene is more realistic and makes common sense, jointly considering the position and orientation?"}
If the reciprocal rank is close to 1 of a method, it indicates the generation result by the method ranked first among the annotators.

\subsection{Compared Approaches}
We compare our method with two state-of-the-art approaches: HoloDeck and AnyHome:

\noindent\textbf{AnyHome}~\cite{AnyHome} leverages an LLM to generate the spatial relationships between objects and then uses a rule-based algorithm to generate a coarse layout. \footnote{Since the official code of AnyHome does not include the refinement stage, we only got coarse results for AnyHome.}
Subsequently, AnyHome utilizes a score sampling distillation process~\cite{SDS,Fantisia3D} to refine the coarse result.

\noindent\textbf{HoloDeck}~\cite{HoloDeck} also produces spatial relationships between objects with an LLM.
It iteratively adds objects to the scene.
When putting each object, it first gets all valid placements without object collisions and 
without exceeding the room’s boundaries.
Then, HoloDeck chooses the optimal place that satisfies most spatial relationships.
The spatial relationships are checked by pre-defined rules.

\subsection{Quantitative Results}
\begin{figure*}
    \centering
    \includegraphics[width=1\linewidth]{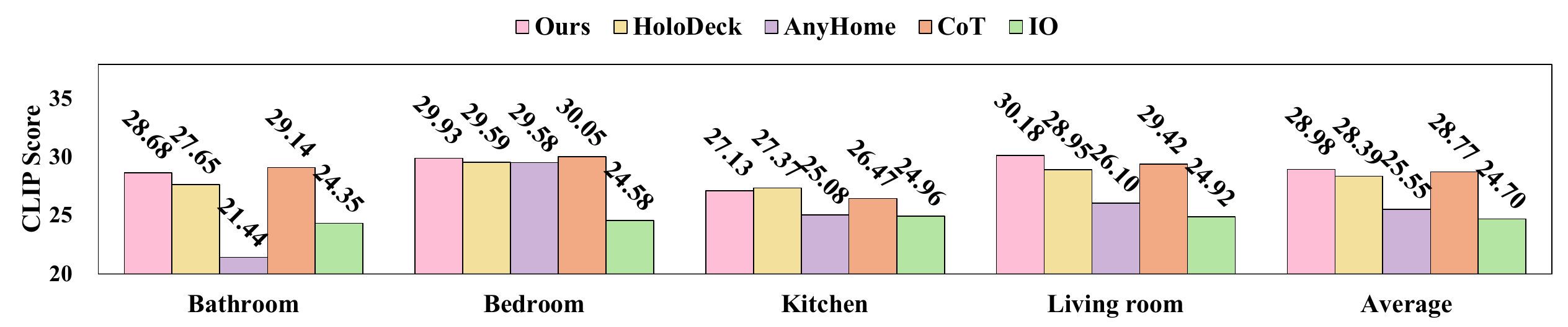}
    \caption{Performance comparison in terms of CLIP score by our proposed model with state-of-the-art methods.}
    \label{fig:clip}
\end{figure*}
As shown in~\Cref{fig:clip}, our proposed method outperforms HoloDeck and AnyHome in most of the scene types.
HoloDeck and AnyHome rely on pre-defined rules for object placement, limiting their flexibility.
In contrast, our method puts an individual object with the local tree search module.
This module decomposes the task into several steps and utilizes a VLM to reason in the top-down view space.
The pre-trained VLM incorporates rich common sense and we only use it to solve such small sub-tasks that the VLM can produce reliable responses. 
Nevertheless, our method has less $0.24$ CLIP score than HoloDeck in the \textit{kitchen}.
This may be because the layout of kitchens is commonly place objects along walls and they do not have obvious spatial relationships between the objects.
Besides, by horizontally comparing the results of our method in different scene types, our method achieves $29.93$ and $30.18$ in the \textit{bedroom} and \textit{living room}, respectively, outperforming others.
This may also be because there are obvious spatial relationships in the \textit{bedroom} and \textit{living room}.
For instance, \textit{a TV stand in front of a sofa}, \textit{a nightstand beside a bed}, \textit{a table in front of a sofa}, \etc.

\subsection{Qualitative Results}
\input{tab/cmp_user}

\begin{figure*}
    \centering
    \includegraphics[width=1\linewidth]{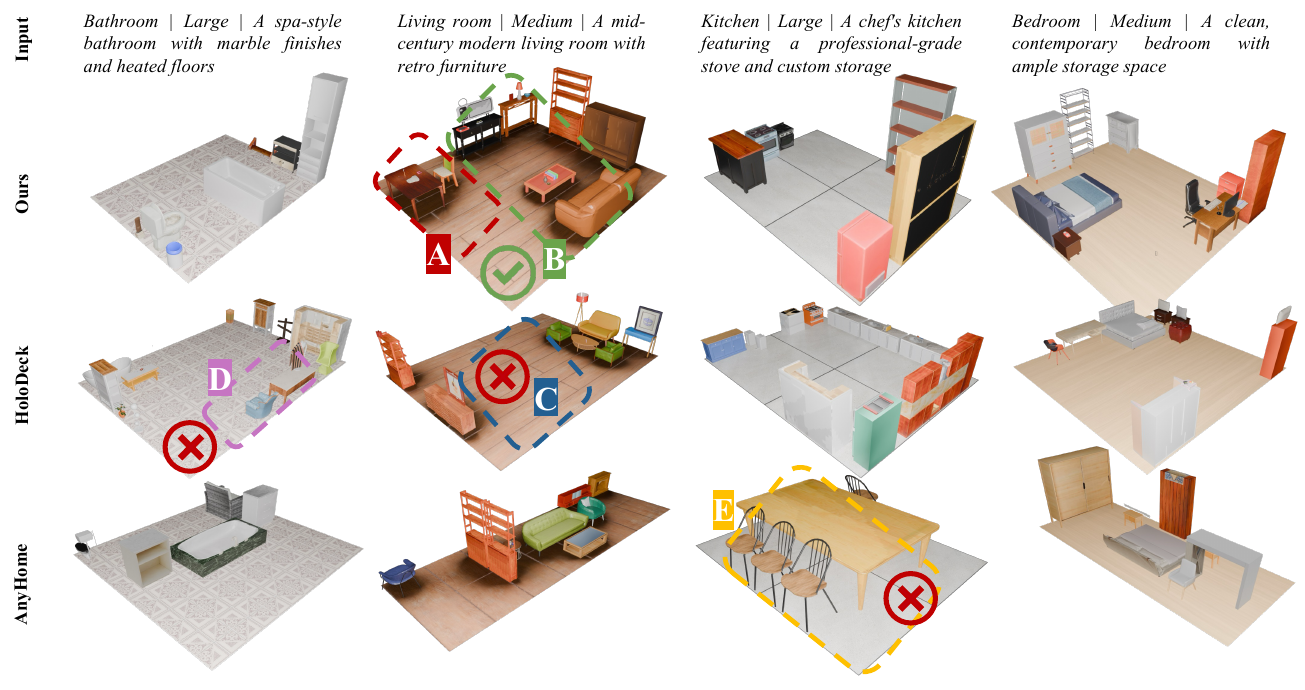}
    \caption{
    \textbf{The generation results for the bathroom, living room, kitchen, and bedroom.}
    For visualization, we manually set floor textures for the rooms.
    The results are rendered with the Blender's Cycles engine.
    }
    \label{fig:visualize}
\end{figure*}

\begin{figure*}
    \centering
    \includegraphics[width=1\linewidth]{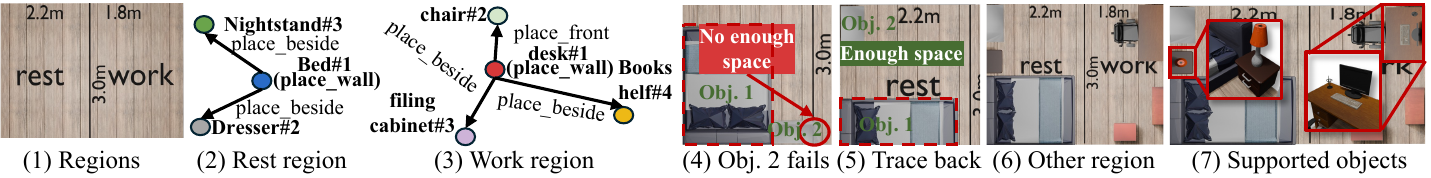}
    \caption{Visualization results of intermediate steps of generating a scene.}
    \label{fig:steps}
\end{figure*}

\begin{figure}
    \centering
    \includegraphics[width=1\linewidth]{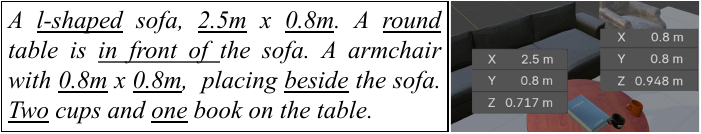}
    \caption{Visualization example of the scene control for complex prompt.}
    \label{fig:control}
\end{figure}

\noindent\textbf{Comparison.}
We show some samples generated by different methods of four scene types in~\Cref{fig:visualize}.
From the visualizations, our method generates more realistic scenes.
For example, objects in the living room generated by our method are well placed.
In~\Cref{fig:visualize} (A), a chair is placed in front of a desk for users to sit on the chair and work on the table.
In~\Cref{fig:visualize} (B), the sofa, coffee table, and TV stand are center-aligned to make users sit on the sofa and watch TV or reach objects on the coffee table.
In contrast, HoloDeck~\cite{HoloDeck} tends to generate objects along walls, leading to large free space in the middle of the rooms in~\Cref{fig:visualize} (C).
Besides, HoloDeck produces a semantically inconsistent result in the \textit{bathroom}, as shown in~\Cref{fig:visualize} (D).
The \textit{armchair} should not be present in a \textit{living room} or bedroom rather than a \textit{bathroom}.
AnyHome's~\cite {AnyHome} results are the worst and produce semantically inconsistent results.
In~\Cref{fig:visualize} (E), the \textit{dining table} and \textit{dining chairs} are not commonly placed in a \textit{kitchen}.
These findings validate the effectiveness of our hierarchical scene representation.
We do not allow a \textit{rest region} present in a \textit{bathroom} or a \textit{dining region} present in a \textit{kitchen}.
Consequently, our hierarchical scene representation can reduce the possibility of providing semantic inconsistency objects.

\noindent\textbf{Intermediate results.}
We show visualization results of intermediate steps in~\Cref{fig:steps}. We can clearly see our method first divides the room into two regions and generate graphs for each region.
In~\Cref{fig:steps} (4), object 2 fails to be put into the rest region because of not enough space.
In~\Cref{fig:steps} (5), we trace back to object 1 and rotate it, enabling to have enough space to put object 2.
In~\Cref{fig:steps} (6-7), we generate the floor objects for each region and supported objects.

\noindent\textbf{Scene control.}
We show a visualized example of the complex prompt input and the corresponding scene in~\Cref{fig:control}.
The generated scene reflects details in the prompt because of the powerful VLM to understand user's instruction.

\noindent\textbf{User study.}
We also conduct user studies to evaluate the generation quality of our method, HoloDeck, and AnyHome. 
We report the mean reciprocal rank of each method in~\Cref{tab:cmp_user}.
It shows our method achieved the most promising results compared with state-of-the-art approaches.
The mean reciprocal rank of our method is $0.793$ on average, which stands for the samples generated by our approach ranking $1/0.793=1.26$ among the annotators.
In addition, our method gets $+0.360$ and $+0.197$ reciprocal rank improvements on average compared with AnyHome and HoloDeck respectively.
Notably, we perform an impressive reciprocal rank score in \textit{bedroom} ($0.846$) and \textit{living room} ($0.868$), which denotes our method has the advantage in such scenes containing obvious spatial relationships. 

\subsection{Ablation Study}
\input{tab/ablation_user}

To validate the effectiveness of our global-local tree search approach in reasoning 3D layout, we conduct studies on the reasoning paradigms on this task.
(1) First, we use a standard Input-Output (IO) method akin to LayoutGPT~\cite{LayoutGPT}.
The IO method uses the room's type, size, and description as input to directly produce the name, location, dimension, and orientation of objects in a scene with an LLM.
(2) Second, we use a Chain-of-Thought (CoT) method~\cite{CoT}.
The CoT method places objects step by step.
If an object fails to be placed in the scene, the algorithm will not adjust previous thoughts.
The IO and CoT approaches are the special cases of our method.
The IO method does not contain multiple intermediate thoughts and the CoT method reasons step by step but does not explore multiple candidates on intermediate thoughts.
We change the maximum attempts $k=1$ for both global and local tree search modules to make our approach degrade to CoT.

We illustrate the results of user studies in~\Cref{tab:ablation_user}.
It shows our reasoning method achieves $0.75$ reciprocal rank on average and outperforms IO and CoT settings with $+0.381$ and $+0.064$ respectively.
In addition,  we also report the CLIP scores on different ablation settings in~\Cref{fig:clip}.
From the figure, our full method outperforms the ablation variants overall.
The standard IO setting performs the worst.
This may be because the training set of the LLM does not contain much 3D layout samples.
Our method first leverages the LLM to produce the hierarchical scene representation.
It serves as a proxy between the input and layout.
Based on this representation, we decompose the layout generation task as small as possible and leverage the VLM to reason in the space.
Consequently, our method gets significant improvement compared with the standard IO setting.

The CoT setting inherits our hierarchical scene representation and task decomposition and also achieves good results.
However, we find a marginal CLIP score improvement in our method compared with the CoT setting (only $+0.21$ improvement on average) and it performs slightly worse than CoT for the bathroom ($-0.46$) and bedroom ($-0.12$) scenes.
The reason may be that we set small $k$s for our full method for effect and cost trade-off.
It leads to our global-local tree search method not fully exploring the problem space.
Besides, the algorithm aims to find a global optimal.
It will prune the sub-tree if it fails to put an object and discards local optimal solutions.
In contrast, the CoT setting will ignore the failure objects, which can lead the algorithm to a local optimal result.

%% file: tab/prompt.tex
\begin{table}[htbp]
\centering
\SetTblrInner{rowsep=1pt}
\begin{tblr}{
colspec={ccc},
colsep={3pt},
rows = {font=\scriptsize},
row{1} = {font={\bfseries \scriptsize}},
column{3}={5.2cm},
cells={halign=c,valign=m},
hline{1,9}={1-3}{1pt, solid},
hline{2}={1-3}{},
}
Type & Size & Description \\
Bathroom & Small &  A cozy bathroom with a compact shower and sleek vanity\\
Bathroom & Medium & A modern bathroom with marble countertops and a spacious shower\\
Bedroom & Medium & A modern bedroom with a comfortable queen-sized bed \\
Kitchen & Medium & A modern kitchen with a kitchen island and stainless-steel finishes\\
Living room & Small &  A snug living room with a rustic coffee table and warm throw blankets\\
Living room & Medium &  A mid-century living room with retro furniture\\
Living room & Large & A living room featuring oversized sofas and a projector setup
\end{tblr}
\caption{
Some example prompts generated by ChatGPT.
}
\label{tab:prompt}
\end{table}

%% file: tab/cmp_user.tex
\begin{table}[htbp]
\centering
\SetTblrInner{rowsep=0.5pt}
\begin{tblr}{
colspec={cccccc},
colsep={3pt},
hline{1,8} = {1pt,solid},
hline{2,4,6} = {0.5pt,solid},
rows = {font=\footnotesize},
row{1} = {font={\bfseries \footnotesize}},
cell{2}{1}={r=2}{},
cell{4}{1}={r=2}{},
cell{6}{1}={r=2}{},
}
Method & Bathroom & Bedroom & Kitchen & Living room & Average \\
AnyHome & 0.421 & 0.422 & 0.528 & 0.401 & 0.443 \\
& $\pm$0.16 & $\pm$0.19 & $\pm$0.26 & $\pm$0.15 & $\pm$0.20 \\
HoloDeck & 0.660  & 0.577 & 0.585 & 0.563 & 0.596 \\
& $\pm$0.27 & $\pm$0.24 & $\pm$0.27 & $\pm$0.23 & $\pm$0.25 \\
Ours & \textbf{0.751} & \textbf{0.834} & \textbf{0.719} & \textbf{0.868} & \textbf{0.793} \\
& $\pm$0.29 & $\pm$0.25 & $\pm$0.29 & $\pm$0.23 & $\pm$0.27
\end{tblr}
\caption{
Mean ($\uparrow$) and STD reciprocal rank of different methods.
}
\label{tab:cmp_user}
\end{table}

%% file: tab/ablation_user.tex
\begin{table}[htbp]
\centering
\SetTblrInner{rowsep=0.5pt}
\begin{tblr}{
colspec={cccccc},
colsep={3pt},
hline{1,8} = {1pt,solid},
hline{2,4,6} = {0.5pt,solid},
rows = {font=\footnotesize},
row{1} = {font={\bfseries \footnotesize}},
cell{2}{1}={r=2}{},
cell{4}{1}={r=2}{},
cell{6}{1}={r=2}{},
}
Method & Bathroom & Bedroom & Kitchen & Living room & Average \\
IO & 0.437 & 0.350  & 0.441 & 0.356 & 0.396 \\
& $\pm$0.23 & $\pm$0.09 & $\pm$0.22 & $\pm$0.11 & $\pm$0.18 \\
CoT & 0.681 & 0.702 & 0.676 & 0.685 & 0.686 \\
& $\pm$0.26 & $\pm$0.25 & $\pm$0.27 & $\pm$0.25 & $\pm$0.26 \\
Ours & \textbf{0.714} & \textbf{0.780}  & \textbf{0.714} & \textbf{0.790} & \textbf{0.750} \\
& $\pm$0.27 & $\pm$0.25 & $\pm$0.27 & $\pm$0.25 & $\pm$0.26
\end{tblr}
\caption{
Mean ($\uparrow$) and STD reciprocal rank of different methods.
}
\label{tab:ablation_user}
\end{table}

%% file: sec/5_conclusion.tex
\section{Conclusion}
\label{sec:conclusion}
In this paper, we proposed a global-local tree search method to boost the reasoning process of VLMs to generate layouts for 3D scenes. To bridge the semantic gap between natural language instructions and 3D scenes, and reduce the search cost, we represented an indoor scene structure hierarchically and incorporated it into the tree of thoughts in VLMs. The extensive experimental results demonstrated that our approach can generate realistic 3D indoor scenes compared with state-of-the-art methods. In the future, we will extend our proposed method into outdoor scenes and AR/VR applications.

%% file: sec/6_ack.tex
\section{Acknowledgment}
This work is partly supported by the Funds for the NSFC Project under Grant 62202063 and U24B20176, Beijing Natural Science Foundation (L243027).